%% file: main.tex
\documentclass[acmsmall,screen,authorversion,nonacm]{acmart}

\usepackage{booktabs}
\usepackage{tabularx}
\usepackage{amsmath}
\usepackage{listings}
\usepackage{lmodern}
\usepackage{tikz}
\usetikzlibrary{shapes.geometric, arrows.meta, positioning, fit, backgrounds}

\AtBeginDocument{%
  }

\definecolor{testimonialgray}{gray}{0.95}
\newcommand{\testimonialbox}[3]{
  \begin{center}
  \colorbox{testimonialgray}{
    \begin{minipage}{0.95\linewidth}
      \smallskip
      \noindent\small\textit{"#1"}
      \smallskip
      \begin{flushright}
        \footnotesize --- \textbf{#2}, \textit{#3}
      \end{flushright}
      \smallskip
    \end{minipage}
  }
  \end{center}
}

\begin{document}

\title{Towards Automating Scientific Review with Google's Paper Assistant Tool}

\author{Rajesh Jayaram}
\email{rkjayaram@google.com}
%\authornote{Equal Contribution}
\affiliation{%
  \institution{Google Research}
  \country{USA}
}

\author{Drew Tyler}
\email{drewtyler@google.com}
\affiliation{%
  \institution{Google Research}
  \country{USA}
}

\author{David Woodruff}
\email{woodruffd@google.com}
\affiliation{%
  \institution{Google Research \& Carnegie Mellon University}
  \country{USA}
}

\author{Corinna Cortes}
\email{corinna@google.com}
\affiliation{%
  \institution{Google Research}
  \country{USA}
}

\author{Yossi Matias}
\email{yossi@google.com}
\affiliation{%
  \institution{Google Research}
  \country{USA}
}

\author{Vahab Mirrokni}
\email{Mirrokni@google.com}
\affiliation{%
  \institution{Google Research}
  \country{USA}
}

\author{Vincent Cohen-Addad}
\email{cohenaddad@google.com}
%\authornotemark[1]
\affiliation{%
  \institution{Google Research}
  \country{USA}
}

\renewcommand{\shortauthors}{Jayaram et al.}

\begin{abstract}
Artificial intelligence is driving a revolution in scientific discovery, accelerating everything from hypothesis generation to mathematical theorem proving. However, this rapid acceleration is creating a systemic challenge: traditional human peer review cannot scale to match the influx of AI-assisted science. Ultimately, to resolve this tension, we must also deploy AI to accelerate the verification and review process itself. 
To frame the discussion around this transition, 
 we propose a taxonomy consisting of four progressive levels of AI-human collaboration in scientific evaluation, and discuss various trade-offs involved with each.
 
As a step toward this future, we introduce the \textbf{Paper Assistant Tool (PAT)}, an agentic AI framework built for deep scientific review and verification. PAT ingests full scientific manuscripts and produces a comprehensive evaluation, checking theoretical results, validating experiments, suggesting improvements, and identifying potential flaws. By utilizing inference scaling techniques, PAT is able to identify deeper issues than a single model call alone, achieving a $34\%$ improvement over zero-shot recall on mathematical errors in the SPOT benchmark. 
Pilot deployments of PAT as a \textit{pre-submission} tool for authors at two major Computer Science conferences---\textbf{STOC} and \textbf{ICML}---demonstrate its ability to identify critical errors and suggest substantive improvements to research papers. By catching errors early, PAT eases the cognitive burden placed on referees, while preserving their control over the outcomes of the review process.

\end{abstract}

%\begin{CCSXML}
%<ccs2012>
% <concept>
%  <concept_id>10010147.10010178</concept_id>
%  <concept_desc>Computing methodologies~Artificial intelligence</concept_desc>
%  <concept_significance>500</concept_significance>
% </concept>
% <concept>
%  <concept_id>10002951.10003317.10003347.10003352</concept_id>
%  <concept_desc>Information systems~Information extraction</concept_desc>
%  <concept_significance>300</concept_significance>
% </concept>
%</ccs2012>
%\end{CCSXML}

%\ccsdesc[500]{Computing methodologies~Artificial intelligence}
%\keywords{Large Language Models, Peer Review}

% ==========================================
% CUSTOM ACMSMALL FOOTER BLOCK
% ==========================================
\authorsaddresses{Corresponding authors: Rajesh Jayaram (\href{mailto:rkjayaram@google.com}{rkjayaram@google.com}) and Vincent Cohen-Addad (\href{mailto:cohenaddad@google.com}{cohenaddad@google.com}).\\
\textcopyright~2026 Google LLC. All rights reserved.}
% ==========================================

\maketitle
\vspace{-0.5cm}

% ==========================================
% GOOGLE RESEARCH LOGO & HEADER RULE OVERLAY (Put immediately after \maketitle)
% ==========================================
\begin{tikzpicture}[remember picture, overlay]
  % 1. Logo aligned with the left title margin (xshift=1.25cm)
  \node[anchor=north west, inner sep=0pt, xshift=1.62cm, yshift=-1.2cm] (logo) at (current page.north west) {
    \includegraphics[height=0.8cm]{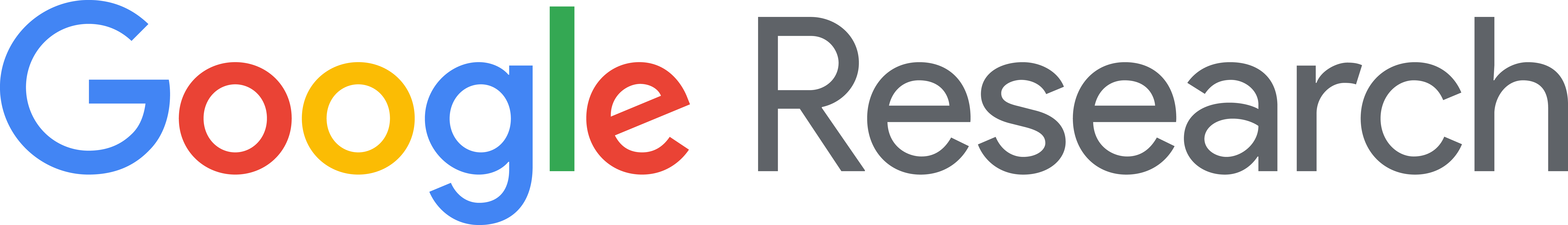}
  };
  
  % 2. Dividing line underneath the logo, spanning the exact text width
  \draw[line width=0.5pt, gray!80] 
    ([yshift=-0.2cm]logo.south west) -- ++(\textwidth, 0);
\end{tikzpicture}
% ==========================================

\section{Introduction: The Scientific Validation Bottleneck}

The rapid advancement of Large Language Models (LLMs) has sparked a revolution in automated scientific discovery.
Models can now generate complex code, infer patterns from massive datasets, and even assist in proving mathematical theorems.
However, the Scientific Method requires rigorously validating these outputs, not just generating them.
As AI-assisted scientific output explodes, the academic community faces the challenge of verifying these results and detecting underlying errors.
While human verification remains the ideal, the cognitive labor required will not scale to keep pace with automated generation.

This validation bottleneck is particularly acute in scientific peer review, which is entirely unequipped to handle the influx of new papers.
In technical fields such as mathematics and theoretical computer science (TCS), comprehensive review requires line-by-line verification of dense proofs, which can take a human reviewer days to accomplish.
Perhaps the most extreme example is found in AI conferences themselves. While submission rates to AI venues were growing prior to 2023 \citep{Yang2025PaperCopilot}, they exploded following the widespread availability of generative AI. As shown in Table~\ref{tab:submission_stats}, the total submissions to three flagship AI conferences have soared over the past three years.
While we cannot definitively attribute this trajectory to AI, there is compelling evidence supporting it: as early as 2024, at least 17.5\% of computer science abstracts on arXiv carried evidence of AI generation~\citep{liang2024mapping}, a figure that reaches up to 40\% in specific biomedical subcorpora~\citep{kobak2025delving}. Undoubtedly, these figures are much higher today.
Given this unprecedented volume, there is significant interest in exploring how AI itself can aid in unburdening the peer review system.

While recent advances in reasoning models have shown remarkable promise in scientific problem-solving, comparatively less work has been done to optimize systems for error detection and manuscript review. 
To address this challenge, we developed the \textbf{Paper Assistant Tool (PAT)}, a verification and review agent that utilizes deep inference scaling techniques to identify flaws and suggest improvements in scientific papers. 
We specialized PAT for mathematics and computer science papers, with the particular aim of addressing the extreme submission rate at AI conferences.

To explore how empowering authors with systems like PAT can help improve the overall quality of papers, we partnered with \textbf{STOC} and \textbf{ICML}---two major computer science conferences---to freely provide PAT to authors prior to submission \cite{stoc2026patblog, jayaram2026icml}. 
The reaction from the community was highly positive. 
In Section \ref{sec:pat-program}, we discuss these pilots and their outcomes. Given the positive community reaction, we plan to continue improving PAT, enabling the tool to take on more influential roles within the peer review process.

Motivated by these developments and our experience with building PAT, we believe it is necessary to bring further structure to the discussion surrounding AI in science.
In Section \ref{sec:levels} we propose a taxonomy of roles for AI integration in peer review, and explore the trade-offs of each approach.
By categorizing these modes of AI usage, we aim to foster a more transparent and informed discussion within the scientific community. This discussion carries significant weight, as future policy changes could shift the power of publication decisions, which have career ramifications, from human experts to AI agents.

\begin{table}[t]
\caption{Submissions to Flagship AI Conferences (2020--2026).}
\label{tab:submission_stats}
\centering
\small
\setlength{\tabcolsep}{8pt}
\begin{tabular}{lccccc}
\toprule
\textbf{Year} & \textbf{ICLR } & \textbf{ICML } & \textbf{NeurIPS} & \textbf{Combined Sum} & \textbf{YoY Increase} \\
\midrule
2020 & 2,594~\citep{Yang2025PaperCopilot} & 4,990~\citep{Yang2025PaperCopilot} & 9,467~\citep{Yang2025PaperCopilot} & 17,051 & -- \\
2021 & 3,014~\citep{Yang2025PaperCopilot} & 5,513~\citep{Yang2025PaperCopilot} & 9,122~\citep{Yang2025PaperCopilot} & 17,649 & \textbf{+3.51\%} \\
2022 & 3,422~\citep{Yang2025PaperCopilot} & 5,630~\citep{Yang2025PaperCopilot} & 10,411~\citep{Yang2025PaperCopilot} & 19,463 & \textbf{+10.28\%} \\
2023 & 4,955~\citep{Yang2025PaperCopilot} & 6,538~\citep{Yang2025PaperCopilot} & 12,345~\citep{Yang2025PaperCopilot} & 23,838 & \textbf{+22.48\%} \\
2024 & 7,304~\citep{Yang2025PaperCopilot} & 9,653~\citep{Yang2025PaperCopilot} & 15,671~\citep{Yang2025PaperCopilot} & 32,628 & \textbf{+36.87\%} \\
2025 & 11,672~\citep{Yang2025PaperCopilot} & 12,107~\citep{Yang2025PaperCopilot} & 21,575~\citep{NeurIPSChairs2025Reflections} & 45,354 & \textbf{+39.00\%} \\
2026 & 19,809~\citep{Yang2025PaperCopilot} & 24,371~\citep{ICML2026XPost} & 29,703~(est.) & 73,883~(est.) & \textbf{+62.90\%}~(est.) \\
\bottomrule
\end{tabular}
\caption*{Note: NeurIPS numbers include main-track only submissions. The 2026 NeurIPS figure is an estimate obtained by applying the 2024–2025 percentage growth rate to 2025. We believe the true number will be higher due to acceleration of the rate of increase in submissions.}
\vspace{-1 em}
\end{table}

\input{PAT}

\input{stoc-and-icml}

\input{AI_Roles}

\input{conclusion}
\bibliographystyle{ACM-Reference-Format}
\bibliography{citations}

\end{document}

%% file: PAT.tex
\section{The Paper Assistant Tool: Automating the Scientific Verification Process}
\label{sec:pat-architecture}

In this section we introduce PAT, an agentic paper reviewing and validation system powered by Gemini Deep Think \cite{deepthink2025}.
PAT utilizes inference-scaling techniques to increase the probability that a critical error is found within a given manuscript.
PAT has been specialized for the detection of mathematical and logical errors, as well as comprehensive feedback and analysis of Computer Science papers.
This specialization is especially motivated by the explosion of new papers in Machine Learning conferences. 
Currently, PAT does not produce subjective assessments or rankings of papers, and instead focuses on identifying objective errors and suggesting potential improvements. 

\paragraph{Design Considerations.}
To motivate our design choices, first we consider basic alternatives.
The obvious baseline is a single inference call to a language model with the full paper given as input.
While simple and not totally ineffective given today's models, this approach is limited by the effective context window of the model to perform deep analysis of the entire paper. 
Verifying complex mathematical claims requires generating a large number of ``thinking tokens,'' which can easily exceed a single model's context capacity.

To ameliorate this, one could call the model many times independently (Pass@k) and attempt to synthesize the outputs into a single report.
This approach has two fundamental issues. 
First, while Pass@k scaling increases recall compared to a single generation, it also significantly degrades precision. 
Models are prone to hallucination; an actor using a review agent must check statements made to ensure their validity.
If a model outputs 10 potential issues per pass, and has a 10\% chance of finding a critical issue per pass, running it 10 times forces a human to sift through $\sim$100 proposed issues just to find one valid, critical error.
Second, Pass@k scaling does not fully address the context limitations: 
without orchestration, independent model calls might spend their context budgets on the same sections of a paper, leaving other sections under-validated.

\paragraph{The PAT Pipeline. }
PAT is designed to specifically resolve both of the prior limitations.
PAT employs a ``segmenter''agent to break each paper into segments, each of which is a set of pages which share a logical theme (e.g., experiments, theory, methodology, etc.). The segments can be overlapping and non-contiguous.
The segmenter agent then dynamically allocates a compute budget to each segment based on the information density and complexity of that segment.
This allows PAT to allocate more compute to the most challenging parts of the paper, and save compute for easier manuscripts.

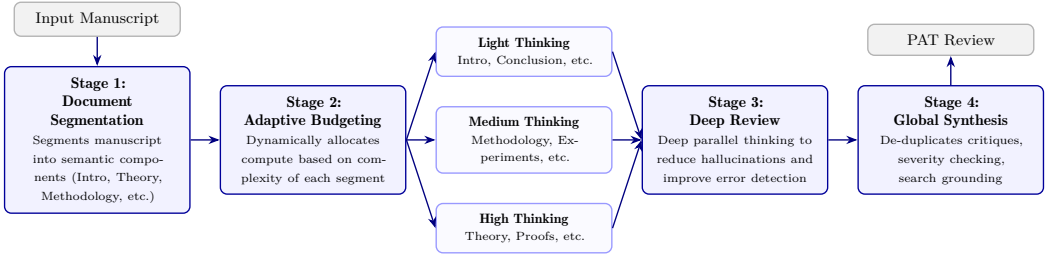
\begin{figure}[t]
\centering
\resizebox{\linewidth}{!}{
\begin{tikzpicture}[
  node distance=0.4cm and 0.6cm,
  base/.style={draw, rounded corners, align=center, font=\small, thick, inner sep=6pt},
  input/.style={base, fill=gray!10, draw=gray!60, text width=3.0cm},
  stage/.style={base, fill=blue!5, draw=blue!60!black, text width=3.4cm},
  branch/.style={base, fill=blue!2, draw=blue!40, font=\footnotesize, text width=3.2cm, inner sep=6pt},
  arrow/.style={->, >=Stealth, thick, draw=blue!50!black}
]

% Nodes
\node[stage] (stage1) {\textbf{Stage 1:}\\ \textbf{Document} \\ \textbf{Segmentation}\\ \scriptsize Segments manuscript into semantic components (Intro, Theory, Methodology, etc.)};

\node[input, above=0.6cm of stage1] (in) {Input Manuscript};

\node[stage, right=0.6cm of stage1] (stage2) {\textbf{Stage 2:}\\ \textbf{Adaptive Budgeting}\\ \scriptsize Dynamically allocates compute based on complexity of each segment};

% Stacked Parallel Thinking Branches (stretched vertically)
\node[branch, right=0.6cm of stage2] (medium) {\textbf{Medium Thinking}\\ \scriptsize Methodology, Experiments, etc.};
\node[branch, above=0.6cm of medium] (light) {\textbf{Light Thinking}\\ \scriptsize Intro, Conclusion, etc.};
\node[branch, below=0.6cm of medium] (deep) {\textbf{High Thinking}\\ \scriptsize Theory, Proofs, etc.};

\node[stage, right=0.6cm of medium] (stage3) {\textbf{Stage 3:}\\ \textbf{Deep Review}\\ \scriptsize Deep parallel thinking to reduce hallucinations and improve error detection};

\node[stage, right=0.6cm of stage3] (stage4) {\textbf{Stage 4:}\\ \textbf{Global Synthesis}\\ \scriptsize De-duplicates critiques, severity checking, search grounding};

\node[input, above=0.6cm of stage4] (out) {PAT Review};

% Arrows (routed horizontally/vertically)
\draw[arrow] (in) -- (stage1);
\draw[arrow] (stage1) -- (stage2);

\draw[arrow] (stage2.east) -- (light.west);
\draw[arrow] (stage2.east) -- (medium.west);
\draw[arrow] (stage2.east) -- (deep.west);

\draw[arrow] (light.east) -- (stage3.west);
\draw[arrow] (medium.east) -- (stage3.west);
\draw[arrow] (deep.east) -- (stage3.west);

\draw[arrow] (stage3) -- (stage4);
\draw[arrow] (stage4) -- (out);

\end{tikzpicture}
}
\caption{System architecture of PAT showing the four-stage inference scaling pipeline. Compute is dynamically partitioned into Light, Medium, and High Thinking tracks based on segment complexity.}
\label{fig:architecture}
\end{figure}

Next, specialized Deep Review agents (powered by either an advanced version of Gemini Deep Think or another proprietary inference-scaling pipeline) verify the contents of each segment.
While each review agent focuses on verifying a subset of the manuscript, it is provided with the full paper as context.
Finally, PAT deploys a synthesis agent to combine the reports from each segment's deep review. 
The synthesis agent utilizes Google search as an extra check for hallucinations (e.g. non-existent papers or theorems), before assembling the final review. 
The flow of the PAT pipeline is illustrated in Figure \ref{fig:architecture}.

This orchestration  explicitly addresses the previously discussed limitations of single-generation and Pass@k scaling. Dividing the paper into logical segments allows models to focus deeply on their assigned section without spending too many thinking tokens on other parts of the paper, making it far more likely that they remain within their context budget.
Further, by utilizing the Deep Review Agent, PAT ensures that the model calls within each segment are coordinated.  This allows for deeper and more efficient review of each section, which is further improved by the dynamic budgeting mechanism. 
Finally, the synthesis agent counteracts the precision degradation typical of Pass@k scaling via grounding and deduplication.

\subsection{Case Study: Verification of Retracted Papers in the SPOT Benchmark}
\label{sec:spot}
One of the primary aims of PAT is to find technical errors in full-length scientific manuscripts. 
Thus, it is useful to attempt to quantify the gains delivered compared with out of the box LLM calls. 
Towards this end, we evaluate the real-world utility of PAT on organic, human-authored errors, using the SPOT benchmark~\cite{son2025ai}, which compiles manuscripts containing verified mistakes that led to subsequent errata or retractions. 
SPOT contains a variety of errors and typos across various literature categories, and is heavily dominated by multi-modal errors such as figure duplication.
In order to isolate clearly verifiable errors and test the deep reasoning benefits of PAT, we filtered the SPOT benchmark for papers in the Mathematics and Computer Science categories containing \textit{"Equation / proof"} errors. 
This yields an evaluation subset of 26 papers with 29 errors.  

For each paper in the benchmark, we run our reviewing pipeline (either PAT or a single model generation) to generate a detailed error report.
Following the evaluation protocol in the SPOT paper, we utilized an automated LLM-based grader to evaluate whether each report contains the ground truth error. While the SPOT paper utilized a strict grader which requires exact keyword matches with the ground truth for a correct score, we use a specialized grader that reasons to determine whether an issue in the reviewer is \textit{logically equivalent} to the ground truth error. Thus, our numbers are not directly comparable to those reported in the SPOT paper.
 To guarantee grading reliability, a human reviewer among the authors audited each grade from the autograder to ensure alignment.

\begin{table}[h]
\caption{Verification Accuracy on the Math/CS Equation and Proof Error Subset of the SPOT Dataset.}
\label{tab:spot_results}
\centering
\small
\begin{tabular}{lc}
\toprule
\textbf{Verification Method} & \textbf{Detection Accuracy (\%)} \\
\midrule
Original SPOT SOTA ~\citep{son2025ai} & 21.1\% \\
%Gemini 2.5 Pro (Zero-Shot) & 48.2\% \\
Gemini 3.1 Pro~\cite{gemini31pro} (Zero-Shot) & 55.2\% \\
%\textbf{PAT (Gemini 2.5 Pro)} & \textbf{73.1\%} \\
\textbf{PAT (Gemini 3.1 Pro)} & \textbf{89.7\%} \\
\bottomrule
\end{tabular}
\end{table}

%\TODO{Incoporate the new eval numbers.}
 
The results of this case study, presented in Table \ref{tab:spot_results}, demonstrate already that modern foundation models, even without inference scaling, perform strongly at the task of scientific error detection, with Gemini 3.1 Pro~\cite{gemini31pro} achieving over $50\%$ error recall over the dataset. This demonstrates a significant leap forward compared to the findings in the original SPOT paper~\citep{son2025ai}, which can be attributed both to the newer generation of models and our logic-aware grader which correctly grades mathematically equivalent descriptions of an error.  This rapid improvement suggests that even zero-shot AI pipelines are now useful scientific tools. 

 PAT, on the other hand, significantly outperforms single model generation, achieves a $34\%$ gain improvement and increasing recall to $89.7\%$. This illustrates the clear performance gains achieved by specialized inference-scaling pipelines.  In particular, zero-shot models are more likely to accept complex mathematical claims without critically scrutinizing them. For example, in a paper on dual Banach spaces \cite{blecher2024missing}, the zero-shot baseline accepted a false complete contractivity claim for real linear maps between complex minimal operator spaces. PAT, however, refused to accept this claim at face value; instead, it constructed a concrete counterexample, exposing a fatal gap that invalidates the main theorem.

This performance carries a strong implication: were arXiv to implement an automated, single LLM call to review each submitted paper, \textit{more than half} of the errors in these retracted papers would have been caught prior to submission. Had the authors utilized PAT, nearly \textit{all} of them would have caught the error.
 Untracked errors propagating through the literature pose a serious problem for the development of science. 
 The performance of models on these error detection tasks, therefore, makes a strong case for their incorporation into multiple stages of the scientific publication process. One could even imagine such a system being directly incorporated into arXiv, allowing for immediate feedback and error identification for submitted manuscripts.

%% file: stoc-and-icml.tex
\section{PAT Experimental Programs at STOC and ICML}
\label{sec:pat-program}

In an effort to empower the academic community with PAT's enhanced reviewing capabilities, we  partnered with several high profile conferences over the last year. Specifically, in November 2025, we partnered with the Symposium on the Theory of Computing (STOC, a premier theoretical computer science conference), and in January 2026 we partnered with the International Conference on Machine Learning (ICML, a premier machine learning conference).\footnote{Announcements can be found in the blog posts \cite{stoc2026patblog,jayaram2026icml}}

For these pilot programs, we provided \textit{only the authors} with access to a PAT review, run once on their manuscript, several days to weeks prior to the final submission deadline. The goal was to provide valuable pre-submission feedback, allowing authors to address potential errors and shortcomings of their work, thereby improving their paper and the overall submission pool. As we will discuss in Section \ref{sec:levels}, the usage of PAT in these pilot programs falls into the category \textbf{Role 1: Tool for Authors}; in other words, PAT was only offered to authors, and not utilized within the formal peer review process.

%%%
The initial deployment of PAT during the STOC 2026 pilot program was optimized specifically for the math-heavy papers submitted to TCS venues. This pipeline focused purely on mathematical rigor, using deep parallel thinking to find logical errors in proofs. For the ICML program, we needed to generalize PAT to handle the significantly more varied forms of manuscripts submitted to machine learning conferences. This expanded system allows PAT to provide detailed critiques of experimental frameworks, spot confounding factors, and point out missing comparisons and experiments. This generalized system is the one presented in Section \ref{sec:pat-architecture}. Both deployments were powered under the hood by an advanced version of Gemini 2.5 Deep Think. Between the two conferences, over \textbf{4,700} submissions were reviewed by the PAT system.

\begin{table}[t]
\caption{Comparative Survey Results: STOC vs. ICML Pilot Programs}
\label{tab:survey_results}
\centering
\begin{tabular}{lcc}
\toprule
\textbf{Survey Question} & \textbf{STOC Cohort ($n=124$)} & \textbf{ICML Cohort ($n=733$)} \\
\midrule
\textbf{Would use PAT Again} & 97\% & 92.1\% \\
\textbf{Improved Paper Clarity or Readability} & 85.1\% & 87.0\% \\
\textbf{Believe PAT has Education Value} & 75.2\% & 83.9\% \\
\textbf{Very or Mostly Helpful} & 92.7 \% & 90.7 \% \\ 
%\textbf{Reasonably Helpful} & 54.5 \% & 68.9 \% \\
%\textbf{Desire Future Access} & 86.2\% & 96.1\% \\
\textbf{Feedback Mostly or All Grounded} & 55.8\% & 64.8\% \\
\textbf{Identified Substantive Theory Gaps} & 11.6\% & 35.4\% \\
\textbf{Ran New Experiments} & -- & 31\% \\

\bottomrule
\end{tabular}
\end{table}

\subsection{Quantitative Author Feedback for PAT Experimental Programs}

Author feedback on the tool for both STOC and ICML was largely positive (Table \ref{tab:survey_results}). For both programs, the vast majority of authors indicated that they would want to use PAT again for future papers, and that PAT improved the clarity and readability of their papers. Similarly, over $90\%$ said that the feedback was either Very or Mostly helpful. Importantly, given that hallucinations and misunderstandings are key failure points of AI-assisted reviewing, we view the fact that over half found the feedback mostly or all grounded (factual) to be a strong positive result. 

However, the two most significant results were responses to the final two questions in Table \ref{tab:survey_results}. We asked authors if PAT identified \textit{significant} errors in  theoretical results, requiring them to spend more than an hour to fix. For STOC, one in ten authors agreed, which is arguably a high statistic given that proofs in submissions to STOC are rarely checked in their entirety, thus it is likely that these errors would have otherwise gone unnoticed. Surprisingly, PAT found significant theory errors in more than one in three ICML respondents' papers. This uptick is likely caused by the fact that ICML is not a theory conference, and therefore submissions may adhere to a lower level of mathematical rigor. Importantly for a machine learning conference, $31\%$ of ICML respondents said that they ran totally \textit{new} experiments as a result of PAT's review---a significant finding given the time investment required of these authors. 
This demonstrates PAT's ability to suggest significant improvements to papers in addition to detecting errors.

\subsection{Qualitative Author Feedback for PAT Experimental Programs}
\label{sec:qualitative}

Further evidence for PAT's efficacy comes from the large swath of qualitative feedback provided by authors. The tool demonstrated a capacity to uncover subtle flaws that frequently elude human expert detection. Several specific cases include:

%While many automated tools excel at superficial formatting, PAT successfully identified fatal mathematical and algorithmic flaws:
\begin{itemize}
    \item \textbf{(The Fatal Algorithm Bug):} \textit{"It found a critical bug in a way we were applying a tool... an embarrassingly simple bug that evaded us for months. Luckily we were able to fix it but it meant we had to completely change and add 7-8 pages worth of technical content."}
    \item \textbf{(Invalid Proof):}  \textit{"Contradiction in the Analysis of the Unbounded Time Regime... The proof was invalid and we fixed that after AI found it. The lemma itself is a simple fact but nevertheless we still need a correct proof."}
  %  \item \textbf{(Subtle Bound Gap):} \textit{"One technical gap is the L-Lipschitz condition I used... It turns out this condition is much stronger than what I needed... Such a technical gap is fixed by the new Definition 3..."}
\end{itemize}
Overall, the qualitative responses to PAT have been highly positive. Authors from the STOC program were generally surprised by the level of mathematical rigor that PAT is able to assess and critique.

\testimonialbox{The [..] set of comments were simply mind-blowing! It pointed out a subtle though fatal bug in my algorithm, which I managed to fix on time!}{Vijay Vazirani, Distinguished Professor}{University of California Irvine}

\testimonialbox{I found the tool very helpful, especially for long papers... the tool caught a pretty significant error, leading to the rewriting of an important claim.}{Hung Le, Associate Professor}{UMass Amherst}

\testimonialbox{It spotted a technical error that was easily fixable but still took me two hours to write down.}{Jason Li, Assistant Professor}{Carnegie Mellon University}

Even for papers that did not have critical errors to detect, PAT frequently found ``mathematically significant typos'', meaning typos or small errors that significantly degrade rigor and readability. Examples included missing absolute value signs, inequalities pointing in the wrong direction, overloaded notation, and off-by-one errors. Feedback from the ICML cohort strongly reinforced PAT's utility for finding such errors. 

Despite these positive outcomes, pilot testing also highlighted several limitations. The most frequently reported challenges were: (1) date hallucinations and outdated knowledge cutoffs, (2) PDF parsing issues, and (3) falsely claiming a proof or argument is incorrect due to failures in reasoning or model misunderstandings. We have addressed first two issues by better search tooling and parsing. The third issue is a product of any LLM based system, and we are actively addressing it via improvements to PAT's reasoning capabilities.

%% file: AI_Roles.tex
\section{Categorization of Roles of AI Automation in Peer Review}
\label{sec:levels}

AI has already begun to play a significant role in both the process of writing and reviewing scientific papers. A recent study by Pangram~\cite{pangram2025iclr} found that 21\% of ICLR 2026 Reviews were \emph{fully} AI-generated, despite being a violation of conference policies. Given the rapid improvement of models and the burden of reviewing papers, this number will surely continue to grow. Thus, the best path forward may be to agree on principled and safe policies for AI usage in peer review.  A positive example in this direction was the AAAI-26 AI Review Pilot~\citep{biswas2026ai}, where a specialized pipeline was deployed to generate a clearly identified LLM review for every main-track submission.  We believe this was only the beginning of what will be a significant change in peer review.

 To aid in discussions around AI usage in peer review, we propose an explicit taxonomy of four ``roles" which AI can enact within this process, and discuss the potential benefits and harms of each. 
Our taxonomy parallels both the SAE Levels of Vehicle Autonomy, as well as the levels of autonomy for mathematical research presented in~\cite{feng2026towards}. For instance, given sufficient evidence that AI tools are better than humans at reviewing scientific papers, would organizers be willing to allow them to make conference acceptance decisions? Such decisions carry hefty consequences for both science and for authors’ personal careers. A primary goal in delineating these roles of AI usage is so that their merits can be assessed against each other.

\paragraph{\normalfont\textbf{Role 1: AI as a Tool for Authors.}}

 In this role, AI is used purely as a tool for authors in the preparation of their final manuscripts before submission to a publication venue. This level encapsulates the usage of PAT in the pilot programs at STOC and ICML. AI tools may catch bugs in the work, suggest improvements, or automate experiments and coding tasks. However, authors ultimately are fully responsible for their work; Role 1 is not considered to be automation of the paper writing process. While utilizing AI in Role 1 may improve the overall rigor of papers by catching bugs early, it may introduce the counter-intuitive drawback of hiding the more obvious issues, making papers look superficially stronger. Consequently, human reviewers may have to exert additional effort to separate the truly strong papers from those which have been well disguised by AI tools.

 \paragraph{\normalfont\textbf{Role 2: AI as a Tool for Reviewers.}}
Here, AI is used by reviewers in a similar capacity as it was used by authors in Role 1. 
AI tools can be called and interacted with to understand the paper, point out potential flaws, and produce draft reviews. However, the final review is again the full responsibility of the human reviewer. 
While Role 2 speeds up the initial stages of reviewing a paper, AI models may introduce hallucinated critiques of the paper, requiring the reviewer to shift their attention from verifying the paper to verifying the AI-generated review.  Additionally, reviewers are still tasked with identifying issues which may not have been found by the AI. 
Without a coherent LLM-usage policy at a given venue, reviewers may hide their usage of AI tools, and double down on incorrect AI-generated points to protect their professional authority during a rebuttal phase. 
Thus, a successful deployment of Role 2 AI requires a thought-out submission policy, which may allow reviewers to use AI tools but also allow authors to flag points in the review as hallucinated and in need of further review.

\paragraph{\normalfont\textbf{Role 3: AI as a Supporting Reviewer.}}
In Role 3, an AI agent generates a full-length review following the same procedures that a human reviewer would. The agent would not see other human reviews prior to submitting its review, and vice versa. 
%Now there are two main components of a review: the first is the objective validation and analysis of a scientific paper, and the second is the subjective assessment of the paper, which includes a recommendation or rating. 
An AI acting in Role 3 provides only an \emph{objective assessment} of the paper, such as validation of proofs or experimental designs, for a human to later review.
%as well as possibly providing semi-subjective assessments such as on writing quality, for a human to later review.
To allow for subjective assessments, we additionally define \textbf{Role 3.5: AI as a Supporting Reviewer with Ratings} where the AI also provides a subjective assessment, such as a rating or acceptance recommendation, of the paper for a human to judge thereafter. 

Employing Role 3 \& 3.5 shifts the role of the human from a reviewer, who makes recommendations, to Area Chair (AC) who makes decisions. In the most extreme instantiation of this role, all reviews would be AI-generated, and a human AC would only study the reviews and decide whether to accept. On the less extreme side, one could have, for instance, only 2 out of 4 reviews generated by AI systems. Either case results in less demand for human review-hours, which will be a tremendously helpful in reviewing the surge of new AI-accelerated submissions. However, this can also significantly increase the chance of hallucinations affecting acceptance decisions. AI reviewers must be assessed in their accuracy when compared to human reviewers, who are also prone to errors and lapses of judgment.

\paragraph{\normalfont\textbf{Role 4: Total AI Automation of Peer Review}}
 In the future, it is plausible that AI review systems could eventually achieve a level of consistency and precision that outperforms traditional human review in practice, opening the door for a more automated process. 
For example, a NeurIPS 2021 experiment \cite{beygelzimer2021neurips}, replicating an earlier study from 2014 \cite{cortes2021inconsistency},  routed $10\%$ of submissions through two independent review committees and found a $23\%$  inconsistency rate in acceptance decisions. Given the overall $22.7\%$ acceptance rate, purely random selection would yield an inconsistency rate of $35\%$. Thus, the fraction of inconsistent human recommendations is closer to the random baseline than it is to 0. Much of the noise was driven by borderline papers (where ACs had lower confidence in the decision); excluding borderline cases, the inconsistency rate drops to 16\%.

To operationalize this concept in a lower-stakes environment, one could imagine an arXiv-like repository (``AIrXiv'') that hosts only papers vetted by a specialized AI agent. Submissions could undergo multiple stages of automated review, incorporating interactive rebuttals and itemized point-by-point resolutions. Papers that survive many rounds of review would earn a high confidence rating within the system. This interactive, automated pipeline would enable rapid, continual improvement of papers concurrent with their evaluation, ultimately elevating the overall quality of the papers on the repository. Furthermore, this framework could establish a novel publication tier---one more prestigious than a standard preprint but less so than a traditional human-reviewed venue---thereby bypassing bottlenecks in the publication system while still providing a measure of achievement to the author.

Aside from the risks inherited from earlier roles, Role 4 may raise less obvious risks such as a reduction in diversity of opinion due to AI reviewers potentially holding only centralized viewpoints, resulting in a dampening of intellectual debate. Such debate is crucial to the academic tradition, especially in non-technical fields such as the humanities, and must be protected when considering more automated systems.

%% file: conclusion.tex
\section{Conclusion and Future Outlook}
\label{sec:conclusion}

The exponential surge in submissions to computer science conferences has pushed the traditional peer review infrastructure to its breaking point. To address this, we introduced the Paper Assistant Tool (PAT), an agentic reviewing pipeline that utilizes inference scaling to detect deep theoretical, logical, and empirical flaws in scientific papers, and suggest substantive improvements. Pilot deployments at major conferences like STOC and ICML demonstrated PAT's ability to uncover critical errors that had evaded human experts, proving that automated verification can meaningfully alleviate the reviewing bottleneck. 

The success of PAT represents a critical milestone in the taxonomy of AI integration within the scientific process. Currently, systems like PAT can operate successfully within \textbf{Role 1 (Tool for Authors)} and \textbf{Role 2 (Tool for Reviewers)}, acting as powerful tools at the disposal of humans.
 By automating logical validation, human reviewers are liberated to focus their limited bandwidth on deep conceptual novelty and elegance. 
 %, AI can act as a mentor by teaching students while analyzing their work.

Moving forward, the academic community must carefully navigate the transition toward higher levels of AI agency. As the underlying agents improve, we anticipate a gradual shift toward \textbf{Role 3 (Supporting Reviewer)}, where AI operates as a technical reviewer under the supervision of human ACs. As these systems further mature, the paradigm may even evolve toward \textbf{Role 4 (Total AI Automation)}, unlocking alternative publication ecosystems like automated repositories that bypass traditional peer review entirely. In this scenario, AI would assume increasing responsibility alongside human academics for upholding the Scientific Method, acting as both an accelerator of scientific generation as well as a validator thereof.  Beyond peer review, we believe these systems can also be influential in education. Specifically, a natural role of \textbf{AI for Students} will likely emerge, where AI agents are deployed as technical mentors for students, further empowering the next generation of scientific minds.

%However, integrating these high-autonomy systems requires caution. 
However, while AI tools may accelerate validation, ultimately the scientific community remains responsible for the integrity of its own work. Thus, as peer review evolves, human editors will become increasingly essential in safeguarding scientific rigor. Additionally, delineating clear lines of accountability—where algorithmic assessment ends and human judgment begins—will be a critical, open challenge for the academic community. Beyond accountability, the academic community must guard against cognitive complacency (e.g. the deskilling of human reviewers) and ensure equitable access to the compute infrastructure required for these tools. Furthermore, automation may introduce systemic vulnerabilities, such as algorithmic biases and the risk of authors adversarially gaming review agents. Only by addressing these challenges can we safeguard the integrity of the peer review process as we transition toward an automated scientific ecosystem.